\documentclass[twoside,a4paper]{article}

\usepackage[textwidth=15cm]{geometry}

\usepackage{fancyhdr}

\usepackage{url}
\usepackage{hyperref}

\usepackage{graphicx}

\usepackage{amsmath,amsmath,amssymb}
\usepackage{bm}

\newcommand{\affil}[1]{$^#1$}

\title{Temporal Disaggregation of the\\ Cumulative Grass Growth}

\author{Thomas Guyet\affil{1},
		Laurent Spillemaecker\affil{2}\\
        Simon Malinowski\affil{3},
        Anne-Isabelle Graux\affil{4}}

\date{ \affil{1}Inria\, Centre de Lyon, France\\
	      thomas.guyet@inria.fr\\ 
	      \affil{2}Université Rennes 1\, ENSAI, Rennes, France\\
    \affil{3}Université Rennes 1/Inria/IRISA, France\\
    \affil{4} PEGASE, INRAE, Institut Agro, 35590, Saint Gilles, France  \\
          anne-isabelle.graux@inrae.fr\\}
\begin{document}
\maketitle

\noindent\makebox[\textwidth][c]{
    \begin{minipage}{.75\textwidth}
Information on the grass growth over a year is essential for some models simulating the use of this resource to feed animals on pasture or at barn with hay or grass silage. 
Unfortunately, this information is rarely available. 
The challenge is to reconstruct grass growth from two sources of information: usual daily climate data (rainfall, radiation, etc.) and cumulative growth over the year. 
We have to be able to capture the effect of seasonal climatic events which are known to distort the growth curve within the year. 
In this paper, we formulate this challenge as a problem of disaggregating the cumulative growth into a time series. 
To address this problem, our method applies time series forecasting using climate information and grass growth from previous time steps. 
Several alternatives of the method are proposed and compared experimentally using a database generated from a grassland process-based model. The results show that our method can accurately reconstruct the time series, independently of the use of the cumulative growth information.
\end{minipage}}

\pagestyle{fancy}

\section{Introduction}
In 2019, grasslands covered 12.7 million hectares in France, i.e. about 44\% of the useful agricultural surface and 20\% of the national land. 
Therefore, grasslands play an important role, especially by providing ecosystem services such as forage production, climate change mitigation through soil carbon storage, biodiversity maintenance, etc.

The forage production service provided by grasslands is intimately linked to the way the grass grows and therefore to the specific conditions of the year in question.
The grass growth depends on various factors: the water and nutrient resources of the soil, the climate (primarily the radiation useful for photosynthesis, the temperature that regulates the functioning of the plants and the rainfall), the management applied by the farmer (mowing, grazing, fertilization) and the grassland vegetation. 
This growth therefore also depends on the growth that took place on the previous days and that contributed to the establishment of this leaf area.

Information about the grass growth during the year is essential to some simulation models of cattle herds that simulate the use of this resource to feed animals on pasture or at barn (with hay or grass silage). 
But information on grass growth is rarely available. 
The challenge is therefore to design a service that estimates this growth over the year, for example by 10-day periods.

Grasslands are used by farmers to feed their animals. 
They can be grazed and/or cut to produce preserved fodder such as hay or grass silage. The total amount of grazed grass as well as the cut grass during the year corresponds to what is called the ``annual valuation'' of the grassland (expressed in tons of dry matter per hectare and per year). 
Information on the annual valuation of French grasslands is often available. 
In this work, we have assumed that annual valuation allows us to estimate the \textit{cumulative annual growth} of the grass (i.e., the total grass growth over a year), and therefore we assume that the cumulative annual growth is available. 
We will discuss the assumption that this information is available in the results.

Our problem is then to disaggregate the cumulative annual grass growth to reconstruct the dynamics of the growth over the year using time series describing the climate of the year.

We propose to approach this problem as a task of forecasting time series. 
Our problem is not to forecast future values of grass growth. Indeed, time series forecasting methods are used to estimate values of a time series, especially as a function of exogenous variables, such as climate. The main contribution of the present work is therefore to adapt these methods to our problem.

\vspace{10pt}

In the remaining of the article, we first present the data used in this study in Section~\ref{sec:data}. 
Section~\ref{sec:method} presents the formalization of the problem as well as the various methods proposed. These methods are then evaluated in Section~\ref{sec:results}. Before concluding, Section~\ref{sec:soa} positions our approach among those of the state of the art.

\section{Data}\label{sec:data}

For learning and evaluating a model for predicting a time series of grass growth, a database was available from simulations by the STICS model~\cite{brisson2003overview}\footnote{STICS Model~: \url{https://www6.paca.inra.fr/stics/}.}, which is a process-based and deterministic crop model.

\begin{figure}[tbp]
    \centering
    \includegraphics[width=.46\textwidth]{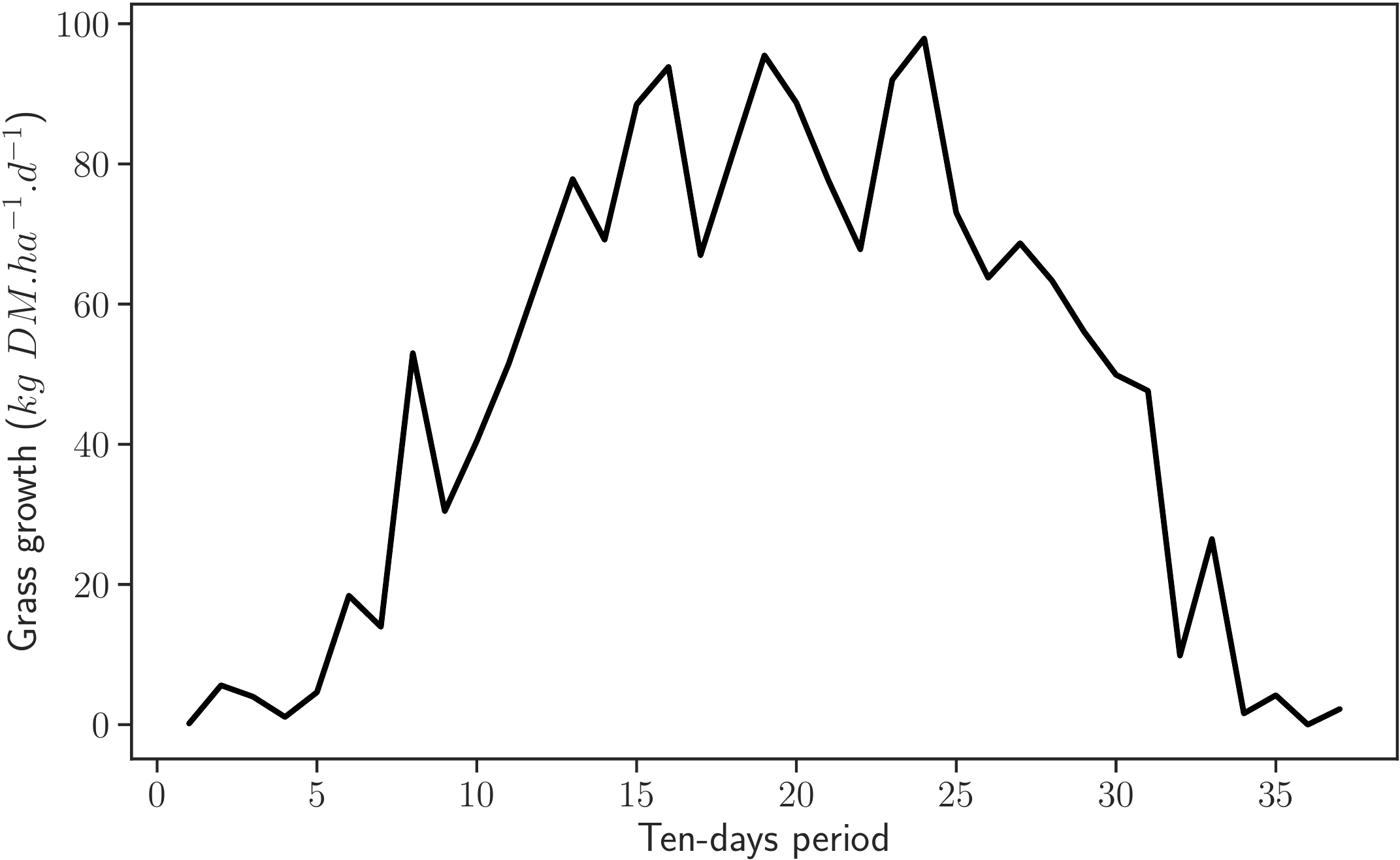} \hfill
    \includegraphics[width=.46\textwidth]{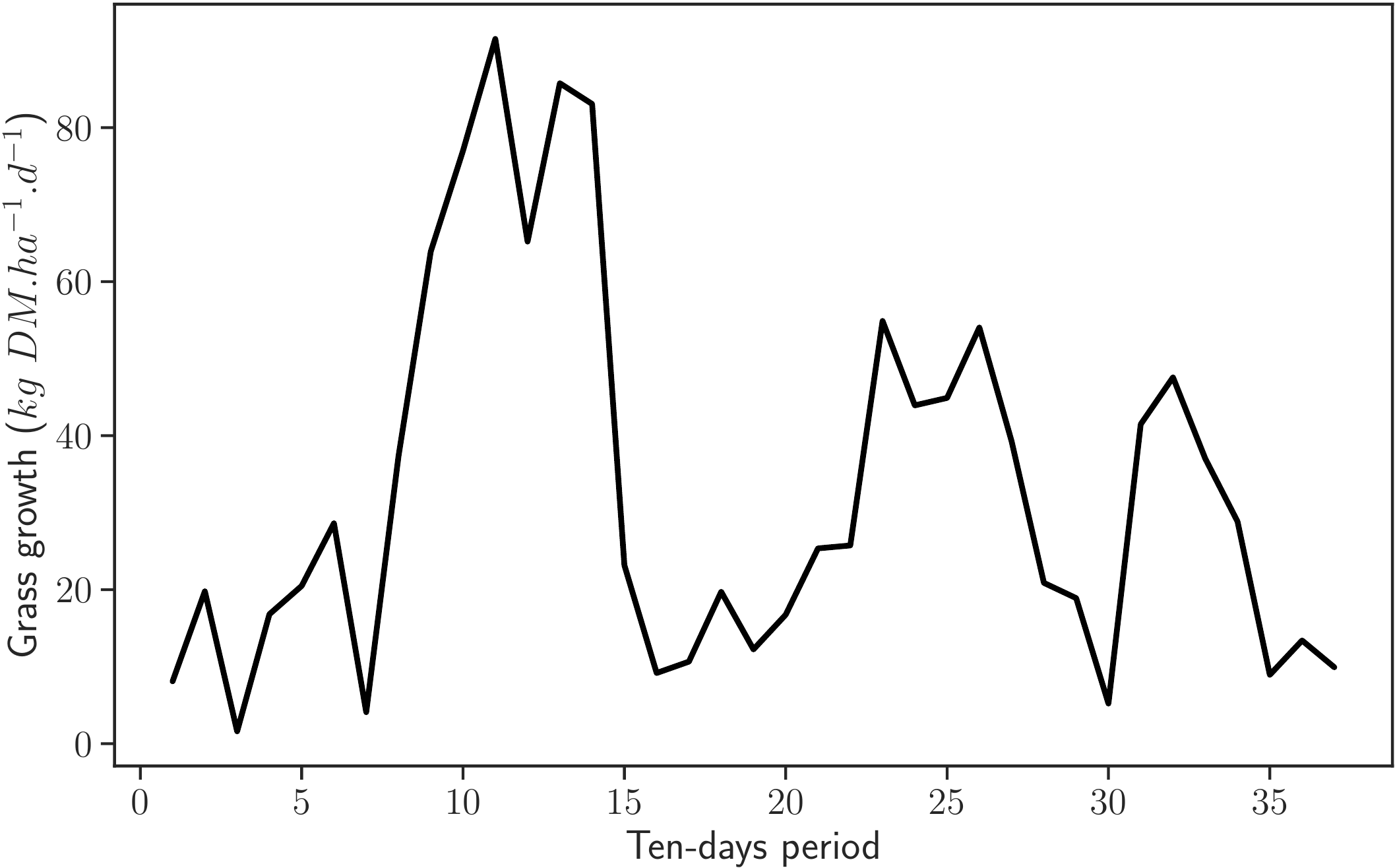}
    \caption{Examples of annual grass growth curve for two different PCUs with different climatic conditions. In the case on the right, there is a sharp decrease in growth in summer, probably due to lack of rainfall.}
    \label{fig:example}
\end{figure}

The simulations were carried out at the scale of France at a high spatial resolution corresponding to pedoclimatic units (PCU), resulting from the crossing of the climatic information (\textit{SAFRAN} grid point) and soil information (soil mapping units), and for which the surface of grassland is significant. The outputs of the simulations correspond to 30-year time series (1984-2013) at daily time step.

Each PCU is associated with a 30-year climate (1984--2013), one to two main soils, one to two main grassland types, and for each grassland type, 1 to 18 farming modes. Within the same PCU, there can be up to 72 grassland simulations. In practice there is an average of ten simulations per PCU. 

Figure~\ref{fig:example} illustrates two examples of grass growth simulations extracted from this database. The two curves have different shapes due to seasonal climatic conditions more or less favorable to the grass growth. 

In the context of this work, we limited ourselves to data from the Brittany region. We selected only the useful variables, namely the daily growth of the grasslands and the daily climatic data. The cumulative growth was obtained by summing the daily growth over a year. We excluded the years of grassland installation because they have a particular grass growth pattern.

The daily growth and climate data were aggregated to the 10-day period (as a sum or an average, see Table~\ref{table_desc_growth_climate}), i.e. over a period of ten consecutive days. Each annual series thus contains 37 values. Agronomists estimated that during 10 days, the growth varies little. Thus, this period length is sufficient for expecting accurate disaggregation. 

The dataset consists of $477~439$ series~\cite{Graux2021:data}. Each series contains 10 variables summarized in Table~\ref{table_desc_growth_climate}: the pair $(id,annee)$ identifies a series of 37 periods of 10 days, the variable measuring growth (to be reconstructed) and the exogenous variables that describe the climate (temperature, rainfall and radiation). The de Martonne index ($im$) is an index of aridity that integrates temperature and rainfall information~\cite{deMartonne26}.
To conduct our experiments, only the last five years (2009 to 2013) were used to reduce computation time. 

\begin{table}[tbp]
\caption{Description of the different variables in the dataset.}
\label{table_desc_growth_climate}
\small
\centering
\begin{tabular}{l c p{8.5cm}}
\hline
Variable~& ~Aggregation~ & Definition (unit) \\ \hline
$id$ & - & Identifier of the simulated PCU  \\
$year$ & - & Simulation year  \\
$period$ & - & Identifier of the period of ten consecutive days \\
$T_{min}$ & min & Minimum temperature ($^{o}C$)  \\
$T_{max}$ & max & Maximum temperature ($^{o}C$)  \\
$T_{avg}$ & avg & Average temperature ($^{o}C$)  \\
$Rain$ & sum & Total rainfall ($mm$)  \\
$RG$ & sum & Global Radiation ($J.cm^{-2}$)  \\
$im$ & - & Martonne index defined by: $I = 37 \times \frac{Rain}{T_{avg}+10}$ ($mm.^{o}C^{-1}$)  \\
$growth$ & avg & Average daily grass growth over the 10-day period ($kg\,DM.ha^{-1}.d^{-1}$, $DM$ meaning Dry Matter) \\ \hline
\end{tabular}
\end{table}

\section{Methodology}\label{sec:method}
In this section, we first formalize the problem of disaggregating a time series as a machine learning problem. 
Then, Section~\ref{sec:data_preprocessing} presents variants of the proposal using alternative representations of a time series (differentiated or cumulative series). 
Finally, Section~\ref{sec:initial_values} discusses the choices of initial values for the reconstructed series.

\subsection{Formalization and General Approach}\label{sec:formalization}
In our disaggregation problem, an example is a couple $\langle C,\bm{Y}\rangle$ where $C\in\mathbb{R}$ is the value of the cumulative growth and $\bm{Y}=\bm{y}_1, \dots,\bm{y}_n$ is a multivariate time series of size $n$ where $\bm{y}_t\in\mathbb{R}^k$ for all $t\in[1,n]$.
$\bm{Y}$ represents the climate data of the year considered.

The objective of the disaggregation is to construct a time series $\widehat{X}=\hat{x}_1,\dots,\hat{x}_n$ in such a way that we have $\sum_{t=1}^{n}\hat{x}_t=C$ and that $\widehat{X}$ is as close as possible to the original time series $X$. 
So we aim to minimize the mean square error (MSE) between $X$ and $\widehat{X}$.


The approach proposed in this work consists in using a time series forecasting technique to reconstruct $\widehat{X}$ step by step. 
The aim is to estimate $\hat{x}_t$ according to the previous values of $\widehat{X}$, but also to the values of the exogenous time series. 

If we consider an autoregressive model of order $p$ ($AR(p)$), $\hat{x}_t$ is obtained by the following equation:
\begin{equation}\label{eq:model_ARX}
    \hat{x}_t = b + \sum_{i=1}^p \varphi_i \hat{x}_{t-i} + \sum_{j=0}^p \bm{\psi}_{j}.\bm{y}_{t-j} = f_{(\varphi,\bm{\psi})}\left(\hat{x}_{t-1,\dots,t-p},\bm{y}_{t,\dots,t-p}\right)
\end{equation}
where $\varphi_i\in\mathbb{R}$, $b\in\mathbb{R}$ and $\bm{\psi}_j\in\mathbb{R}^k$ are the model parameters. 
The order of the model, $p$, designates the number of past values that are taken into account for the prediction.

Note that Equation~\ref{eq:model_ARX} only considers time series values strictly before $t$ to predict $\hat{x}_t$, but since the exogenous values are known, the exogenous data of $\bm{Y}$ at date $t$ is used in the prediction.

More generally, we wish to estimate $\hat{x}_t$ as a function of $\hat{x}_{t-1}, \dots, \hat{x}_{t-p}$, and we note $f_\theta$ this estimation function where $\theta$ represents the parameters of this function. In the experiments, we compared three classes of functions: linear regression, support vector regression (SVR) and random forests (RF).


Our disaggregation problem turns out to be a machine learning problem. We have to estimate the parameters $\theta$ of a forecasting model $f_\theta$ from a training dataset. 
A time series $\widehat{X}$ is then reconstructed by recursively applying this forecasting model. 
Nevertheless, it is necessary to add assumptions about the initial values of the time series to apply the forecasting model the first time (see Section~\ref{sec:initial_values}).

\subsection{Pre and Post Processings}\label{sec:data_preprocessing}
Three pre-processings of the time series have been proposed: no-preprocessing (raw time series), ``differentiated''  time series and ``cumulative'' time series. In the two latter cases, we transform $X$ without modifying the exogenous time series ($\bm{Y}$).

The time series differentiation subtracts the growth for the $(d-1)$-th 10-day period from the growth of the $d$-th 10-day period. 
The learning problem is then to be able to reconstruct accurately the derivative of the growth, rather than the growth itself. 
The reconstruction of the growth time series is done step by step. The value of the differentiated time series must first be predicted and then integrated (cumulated with the previous values) to reconstruct the growth time series. 
The interest of this approach is to leave the choice of the initialization of the growth free during the integration step. 
Thus, the reconstruction of the growth ensures to have a cumulative values equals to $C$ (see post-processings). 

The time series cumulation sums all the values of the previous periods of 10 days. 
This is the inverse path of time series differentiation. 
Our forecasting problem thus becomes to reconstruct accurately this partial cumulative growth. 
During the reconstruction phase, the learned forecasting model reconstructs the cumulative time series which is then derived to obtain the growth time series.
The interest of the cumulation is to potentially ease the learning problem since the predicted function is monotonic.

In addition, three post-processings have been proposed: no post-processing, rescaling and translation. The objective of a post-processing is to enforce the cumulative of the reconstructed time series to equal exactly the known cumulative value $C$. 
Rescaling consists in applying the scale factor $\frac{C}{\sum_{i=1}^n\hat{x}_i}$ to all values of the time series obtained by the prediction model. 
Translation consists in adding the value $C - \sum_{i=1}^n\hat{x}_i$ to all values of the time series obtained by the prediction model. 

\subsection{Order of Models and Initialization} \label{sec:initial_values}
The next step is the choice of the parameter $p$, i.e., the number of past values to be taken into account to predict the next one. Then, the first $p$ values of the growth must be provided to initiate the disaggregation. 
This parameter must be chosen so that it is large enough to give an accurate prediction and, at the same time, it is as small as possible to not require the use of too much \textit{a priori} initial values. 
Domain experts decided to set $p=3$, which corresponds approximately to a month (30 days), and to initiate the reconstruction from the January, a period during which the growth is small and does not changes too much. 

\begin{table}[tbp]
\caption{Initializations of the first 3 values for the preprocessing of a series $X=\langle x_1,\dots,x_n\rangle$.}
\label{tab:init}
\small
\centering
\begin{tabular}{lccc}
\hline
& & Type of preprocessing & \\
Initialization & Raw & Differentiated & cumulative\\ \hline
Concrete & $x_1,x_2,x_3$ & $x_1,x_2-x_1,x_3-x_2$ & $x_1,x_1+x_2,x_1+x_2+x_3$ \\
Average & $9,9,9$ & $0,0,0$& $9,18,27$ \\ \hline
\end{tabular}
\end{table}

Two different initializations of the growth are possible using either the raw time series, or an \textit{a priori} value. 
In this second case, we choose to set a value that corresponds to the average of the concrete growth over each of the first three periods of 10 days and for all the available series ($9\,kg~DM.ha^{-1}.d^{-1}$). Note that the first type of initialization requires information that is not always available in practice, so we would like to compare the second type of initialization with this optimal case.

For the cumulative and differentiated time series, the initialization values are deduced from the previous assumptions. The choices for the initializations are summarized in Table~\ref{tab:init}.

\section{Experiments and Results}\label{sec:results}

The methods were implemented in Python. 
Three regression models were investigated: linear regression (\textit{lm}), Support Vector Regression (SVR)~\cite{awad2015support} (Gaussian kernel, configured with $C=100$) and random forests (RF)~\cite{breiman2001random} (limited to 100 trees). 
We choose the hyper-parameters to make a trade-off between accuracy and efficiency. It has been evaluated on small subsets of the data. 
Other hyper-parameters of the methods are the default ones. 
Due to the large amount of data, the training of the random forests and the SVRs, required to sample the dataset using only the last 5 years of data (total amount of 76~918 series).

Combining the different types of regression and preprocessing yields 9 different growth prediction models:
\begin{itemize}
\item three preprocessings: raw, differentiated (\textit{diff}), and cumulative (\textit{cumul});
\item three regression models: \textit{lm}, SVR and \textit{RF};
\end{itemize}

These models are then combined with several possibilities for disaggregation: 
\begin{itemize}
\item two initializations: concrete or average initialization (at $9\,kg~DM.ha^{-1}.d^{-1}$) of the first three values;
\item use of two post-processings to match the sum of values of the disaggregate time series with the cumulative annual growth $C$: scale factor (\textit{scale}) or translation (\textit{trans}).
\end{itemize} 

Each of these models was learned on the training set and then evaluated on the test set data. 
The training set represents 70\% of the sampled set (30 \% for the test set). Since we wish to have a model that generalizes well for different exogenous data, we choose to split the dataset according to climate: the climate time series of the test set are different from all the climate time series in the learning set. We used \textit{SAFRAN} grid points to split the dataset: the time series of 328 grid points are used for learning, and the time series of the remaining 141 grid points are used for testing.

The accuracy of the time series disaggregation is evaluated by the root mean square error (RMSE). The smaller the better.

As a baseline, we propose a naive model which generates always the average time series. 
This time series has been computed as the average of 10-day growth values for each time series on the training dataset. 
The average RMSE of this naive model is $20.6\,kg~DM.ha^{-1}.d^{-1}$. Then, we expect to lower this error with our proposal.

In the rest of this section, we start by giving the results of the comparisons of our nine models with an average initialization. 
In a second step, we analyze the error committed by the model due to the choice of the initialization, then we analyze the accuracy improvement obtained by using the information of the cumulative growth $C$. 
Finally, Section~\ref{sec:results:quali} illustrates qualitatively the results of the disaggregation.

\subsection{Comparison of the Prediction Models}\label{sec:results:bestmodel}

In this section, we compare our nine forecasting models (three pre-processed data and three types of regressor) and the naive model. In these experiments, time series were initialized with the average initial values and no post-processing adjustment were applied. 

\begin{figure}[tbp]
    \centering
    \includegraphics[width=0.48\textwidth]{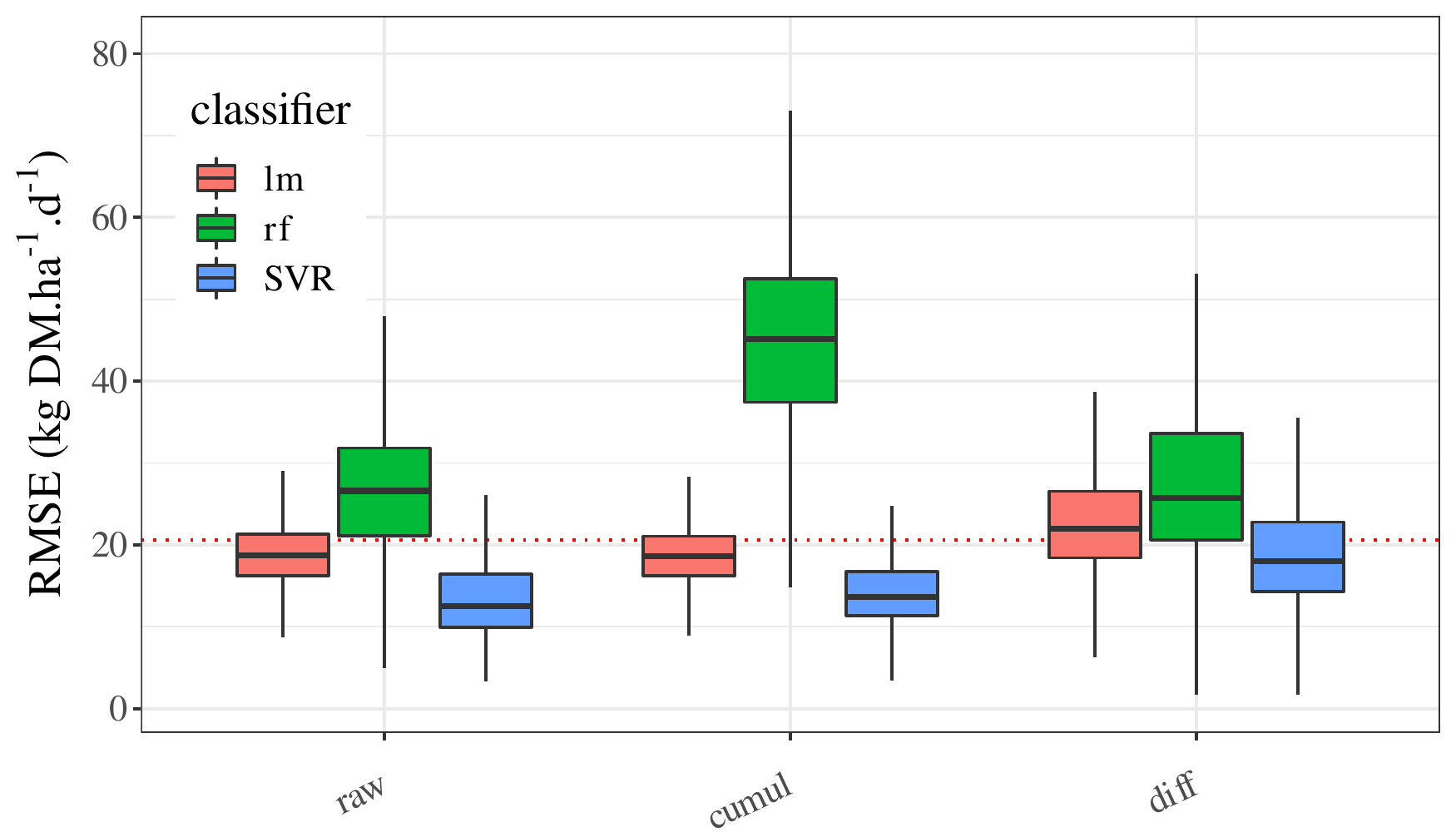}\hfill
    \includegraphics[width=0.48\textwidth]{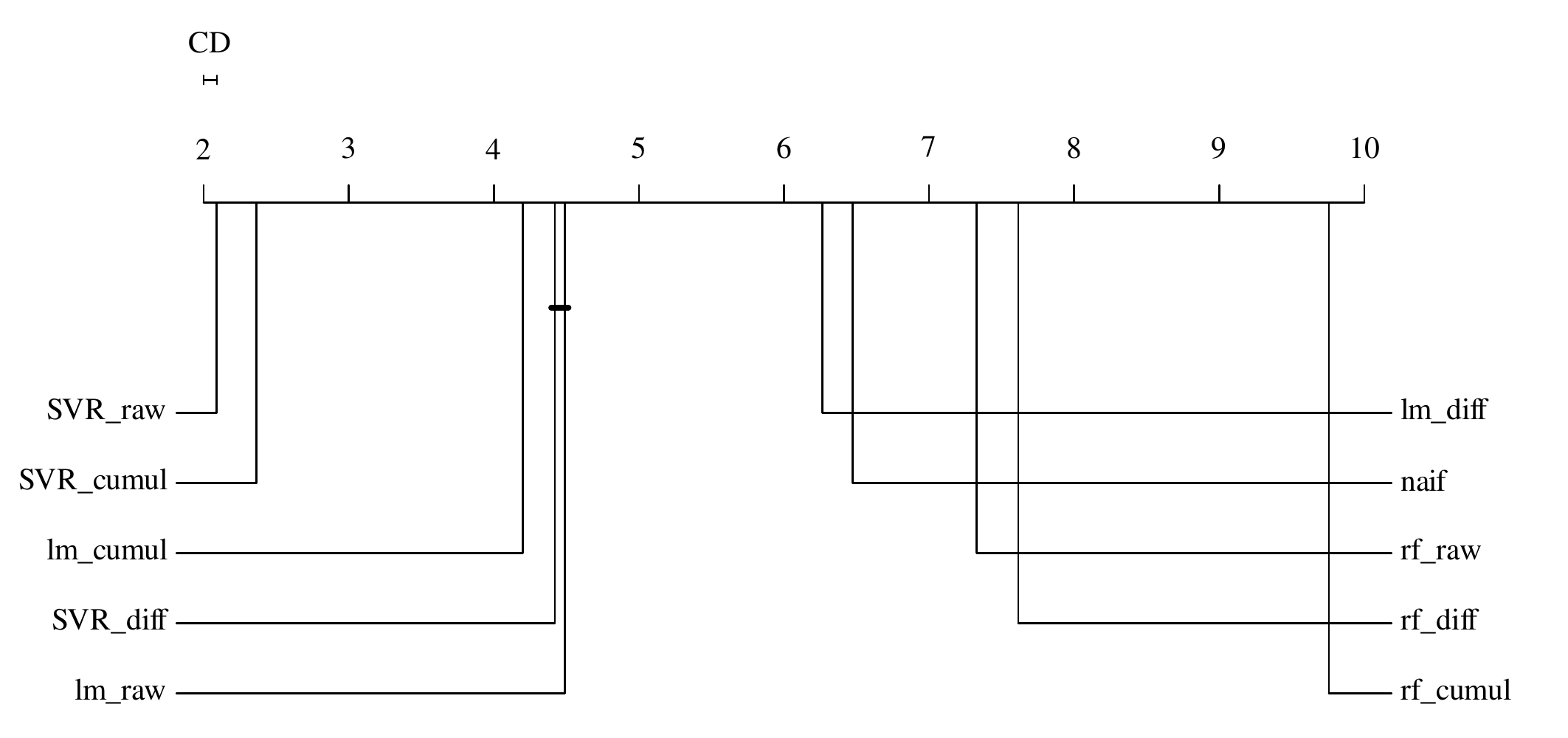}
    \caption{Comparison of RMSE between different classifiers and preprocessings. On the left: error distributions; on the right: critical difference diagram. The dashed line indicates the RMSE of the naive model.}
    \label{fig:cmp_ml}
\end{figure}

Figure~\ref{fig:cmp_ml} on the left represents the RMSE distributions. For the RF, SVR and \textit{lm} classifiers (all pre-treatments combined), the average RMSE are respectively $35.3\,kg~DM.ha^{-1}.d^{-1}$, $16.8\,kg~DM.ha^{-1}.d^{-1}$ and $20.7\,kg~DM.ha^{-1}.d^{-1}$ for the raw time series.  
The better results are achieved with SVR. The random forest model shows surprisingly poor performance compared to the other approaches. It may comes from a model overfitting due to the default hyper-parameters of the RF which iw not necessarily suitable for regression tasks. 
Compared to the average model, only SVR performs better on average. 

The cumulative time series have rather reduced RMSE for the linear (\textit{lm}) and SVR models, on the contrary the RF performance worsens in this case. 

Figure~\ref{fig:cmp_ml} on the right illustrates the same results in a synthetic way by a critical difference diagram.
The diagram confirms that the approaches with SVR are significantly better. The solutions based on cumulation are also rather interesting (except when combined with RF). 
Most of the models based on linear regression and SVR models are significantly better than the naive model. This comparison is based on peer-to-peer differences for each series to be disaggregated (Nemenyi test with $\alpha=5\%$). 

\subsection{Effect of Approximating the First Values}
\begin{figure}[tbp]
    \centering
    \includegraphics[width=.8\textwidth]{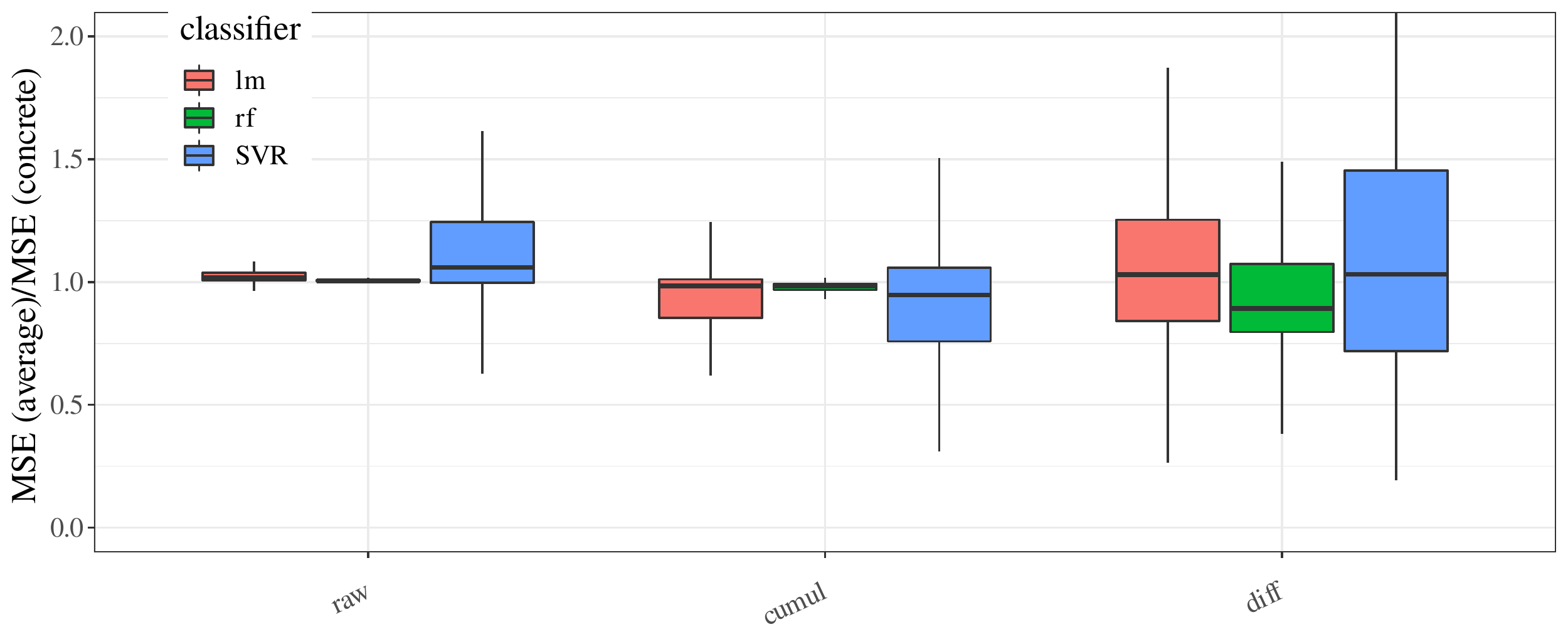}
    \caption{Ratio of RMSE with an average initialisation to RMSE with concrete initialisation. A value greater than 1 indicates that using an average growth value for the first three 10-day periods to initialize the time series is worse than using concrete growth values, and vice versa.}
    \label{fig:cmp_approx}
\end{figure}

This section investigates the errors related to the initialisation of the first grass growth values. It addresses the following question: has the approximation by an a priori average grass growth value an impact on the accuracy of the model prediction? 
Figure~\ref{fig:cmp_approx} visualizes the ratios of RMSE with average initialisation and MSE with concrete initialisation. 

We observe that on average the ratios are very close to $1$. 
This means that the use of the average initialization does not introduce significant error. 
Nevertheless, we note a strong presence of outliers, in particular when using the cumulative or the differentiated time series. These outliers are also observed when using the SVR on the raw time series. 

Finally, we can also see that in some cases, the use of the average initialisation improves the model predictions (ratio $<1$).

\subsection{Improvements by Post-Processings}
This section investigates the accuracy improvement by applying a post-processing. Indeed, the post-processing makes benefit from an additional information: the cumulative growth $C$. Then we expect to lower the RMSE.

\begin{figure}[tbp]
    \centering
    \includegraphics[width=.85\textwidth]{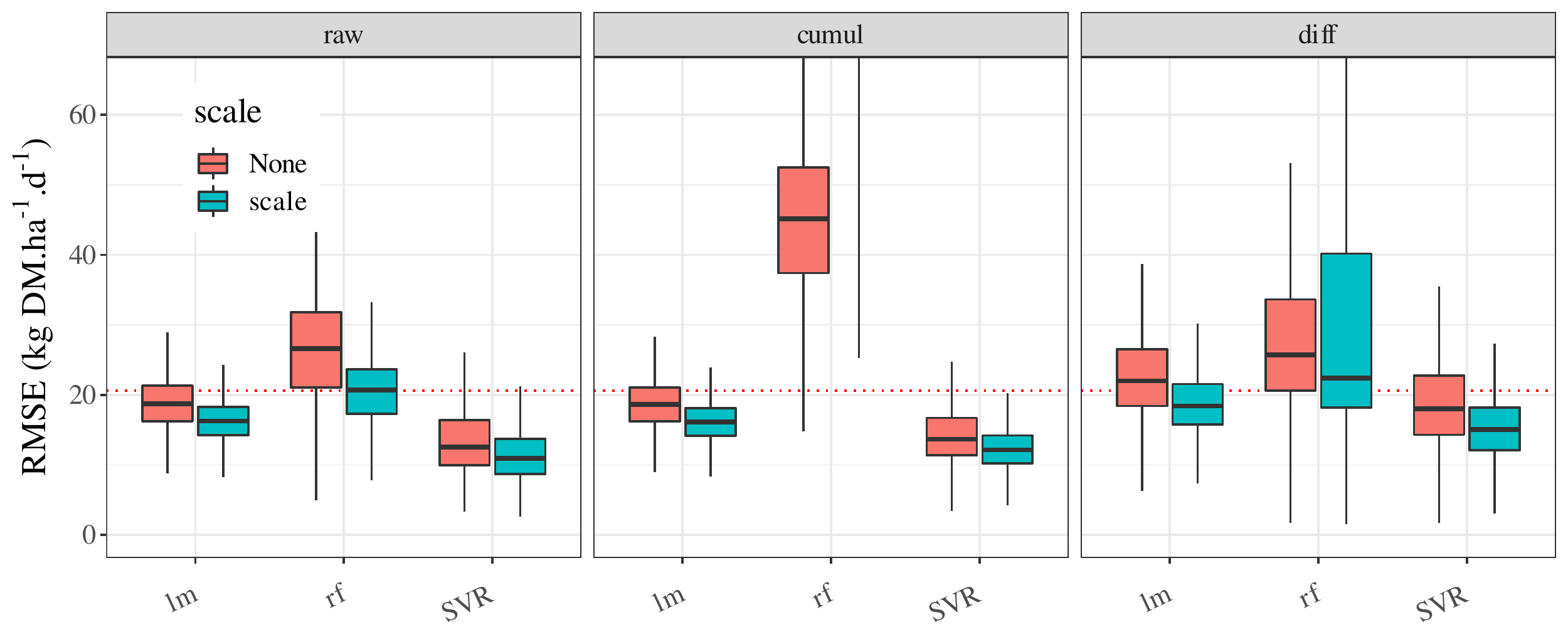}
    \caption{Comparison of RMSE with and without the use of a scaling factor in post-processing (use of the cumulative growth $C$). The box of RF with scaling factor applied on the cumulative time series can not be show on the graph to keep it readable. The dashed line indicates the RMSE of the naive model.}
    \label{fig:cmp_scale_mse}
\end{figure}

We start by looking at the errors while applying the scale factor. 
We can see from the graph in Figure~\ref{fig:cmp_scale_mse} that this post-processing improves the RMSE in most settings. At the median, for the SVR applied without preprocessing, the improvement is $\approx 84\%$. 
The results of RF with cumulative time series continues to contrast with the others. 

Finally, we complement the analysis with the comparisons of the three alternatives of post-processing: translation, scaling factor or without post-processing. This comparison is meaningful only in the case of the differentiated time series. 

\begin{figure}[tbp]
    \centering
    \includegraphics[width=.85\textwidth]{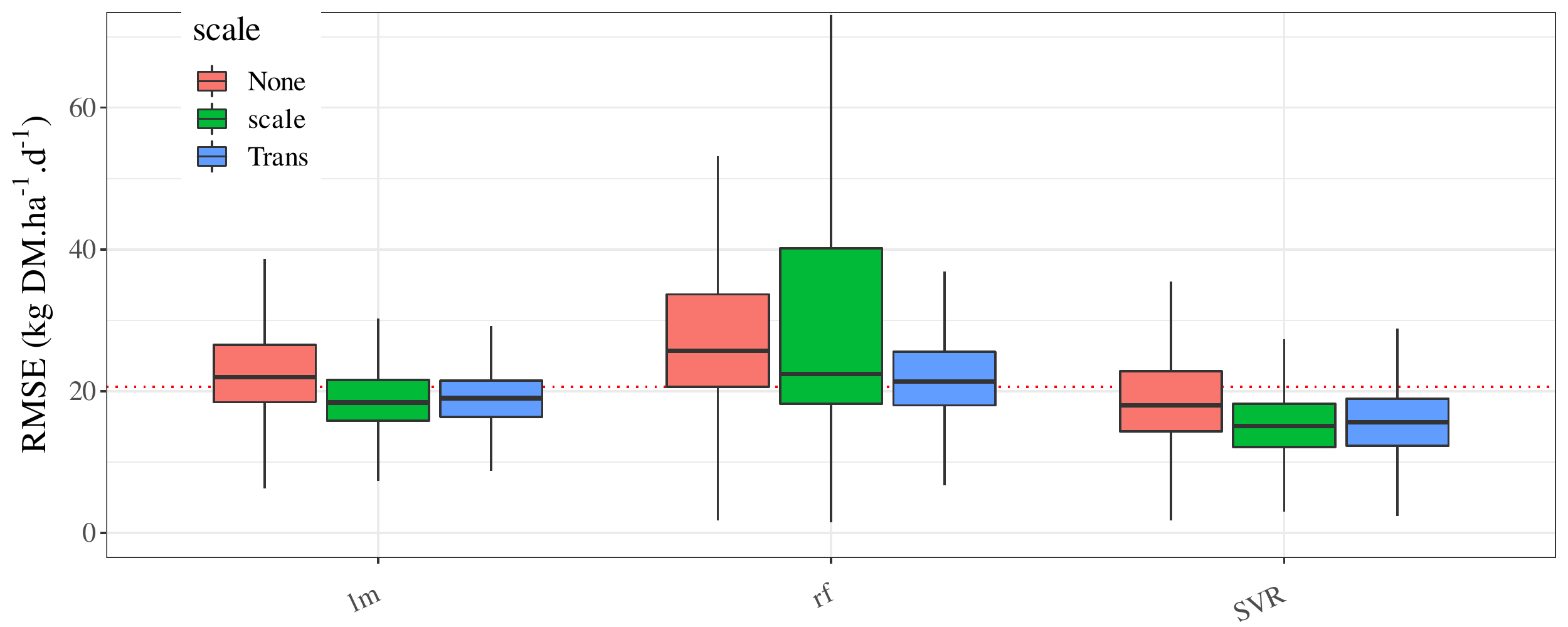}
    \caption{Comparison of the RMSE according to the post-processing applied to the differentiated series: without post-processing, scaling or translation of the curve (using the cumulative growth  $C$). The dashed line indicates the RMSE of the naive model.}
    \label{fig:cmp_scaletrans_mse}
\end{figure}

The results are given in Figure~\ref{fig:cmp_scaletrans_mse}.
Compared to the previous graph, this figure adds the case of the translation. We observe that the performances of scaling are similar to that of translation.
But, translation can induce negative growth values. Such values are not acceptable from a biological point of view. 
Therefore, although attractive from a methodological point of view, the solution with a translation post-processing turns out to be not usable in practice. 
Furthermore, the practical use of scaling tends to show an interest in improving disaggregation by using knowledge of the annual aggregate.

\vspace{10pt}

From these experiments, we can conclude that the best disaggregation solution is the one based on a SVR classifier, without preprocessing the growth series ($SVR\_raw$). 
The average error of this model is $12.4\pm 4.3\, kg~DM.ha^{-1}.d^{-1}$. 
This disaggregation solution can be used with an average initialization of the first three 10-day periods. 
The use of the annual cumulative growth ($C$) is interesting to reduce the RMSE, but the improvement remains weak. 
Remember that this annual cumulative growth is not directly accessible: in the long run, it could be estimated from the annual valuation of grasslands, known from agricultural statistics or interviews with farmers, but the error resulting from this estimation could mitigate the benefit of using this information.

It should be noted that if SVR requires a long computation time (several hours in learning and about ten seconds in inference), learning a linear model takes only a few seconds.

\subsection{Qualitative Evaluation of Reconstructions}\label{sec:results:quali}

\begin{figure}[tbp]
    \centering
    \includegraphics[width=.48\textwidth]{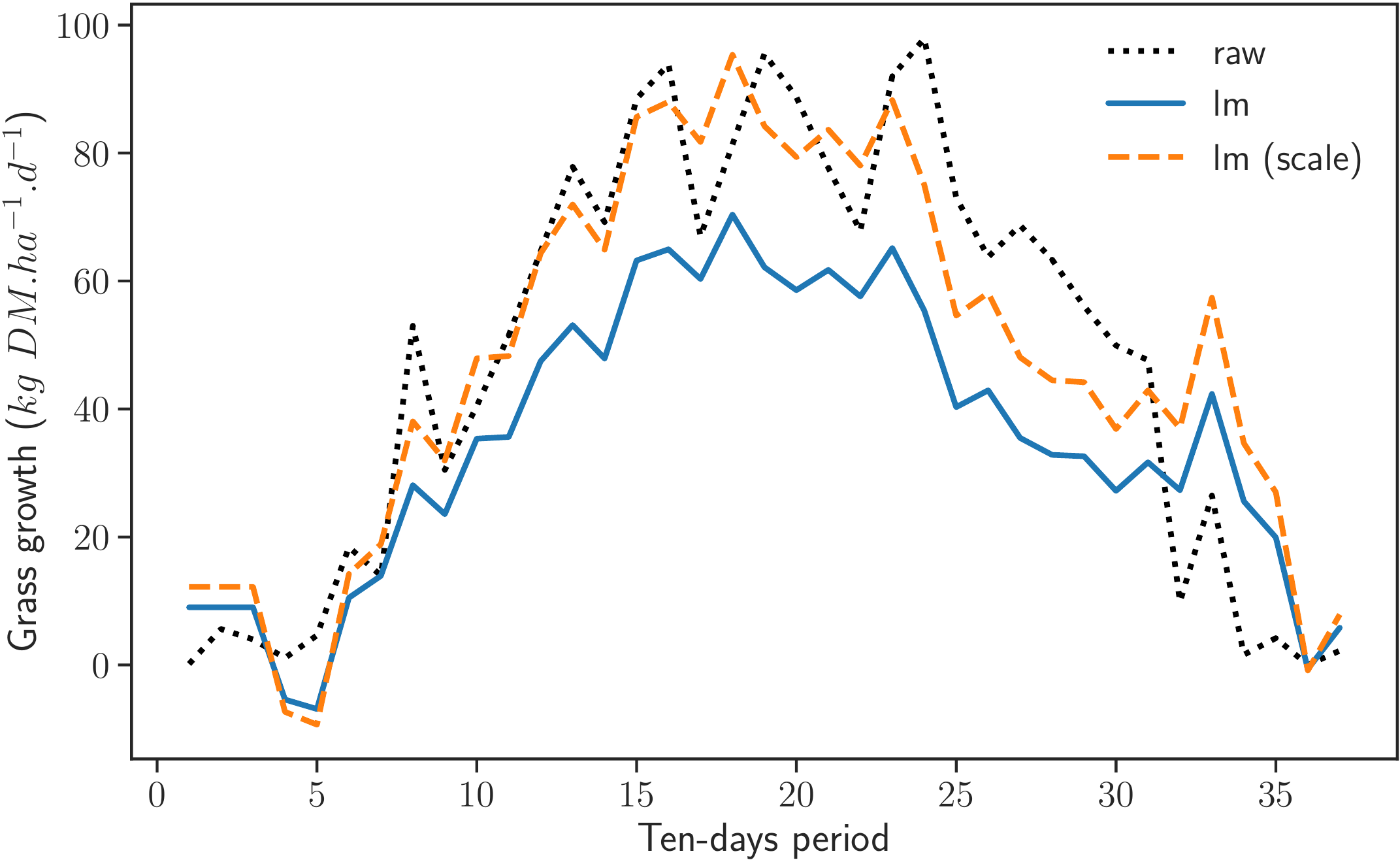}\hfill
    \includegraphics[width=.48\textwidth]{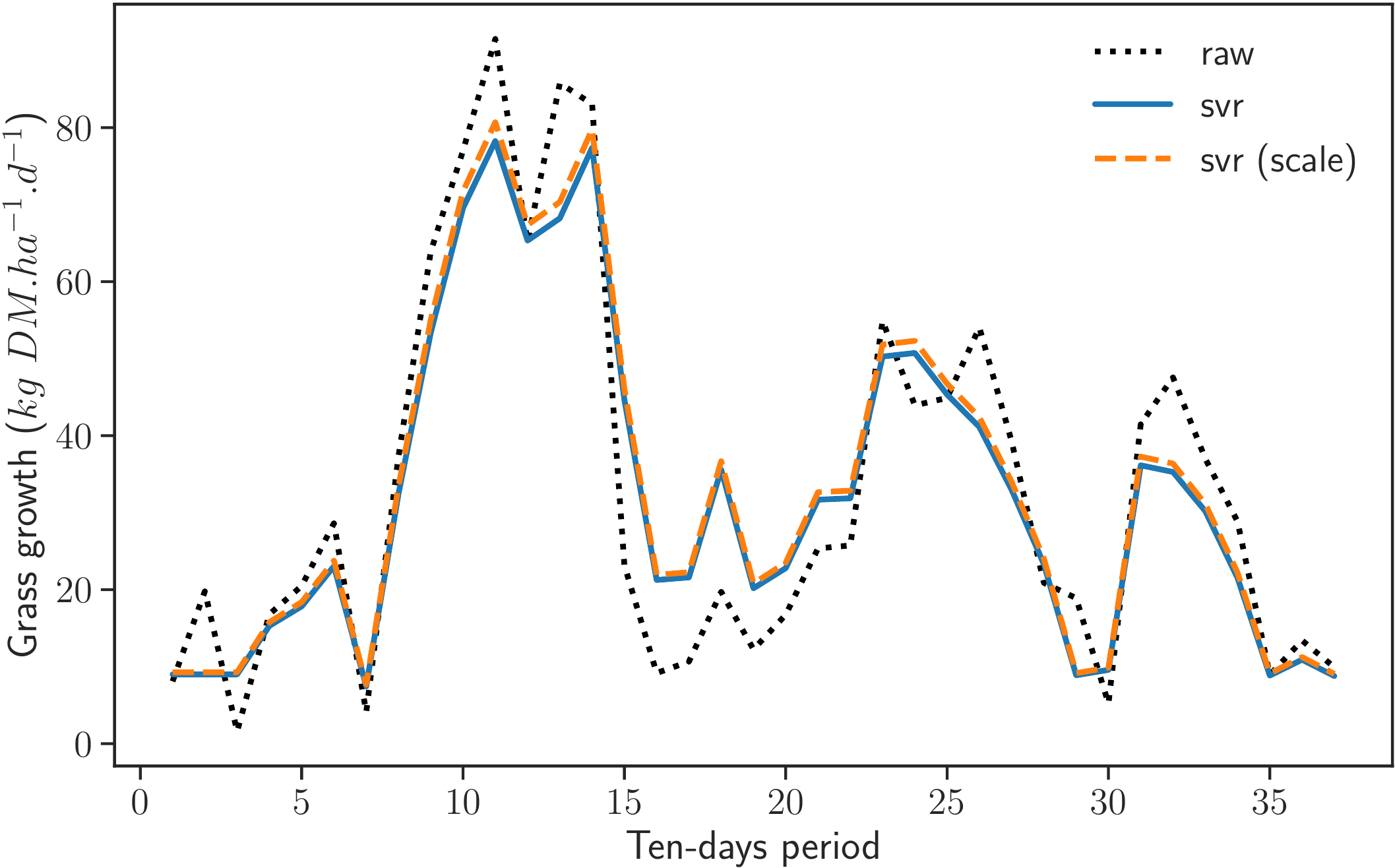}
    \caption{Examples of disaggregation of cumulative growth with \textit{lm} (on the left) and SVR (on the right, RMSE: $6.9~kg~DM.ha^{-1}.d^{-1}$). The example on the left illustrates the effect of post-processing (RMSE before: $17.5~kg~DM.ha^{-1}.d^{-1}$, after: $11.1~kg~DM.ha^{-1}.d^{-1}$). }
    \label{fig:example_desagreg}
\end{figure}

Figure~\ref{fig:example_desagreg} illustrates the disaggregation results obtained for the two time series of Figure~\ref{fig:example}.\footnote{Disaggregation can be tested on-line: \url{https://disaggregation.herokuapp.com}.} 
In both cases, the treatments were carried out without pre-processing and with an average initialization (one can note that the first three values are constant). 
The figure shows that the linear model (\textit{lm}) underestimates growth in summer. The scaling factor corrects this defect, but induces additional errors in winter. On the right curves, we see that the original time series is very well reconstructed despite some important changes. In this case, the linear model does not fit well to these changes.

\section{Related Work} \label{sec:soa}
The term \textit{disaggregation} sometimes corresponds to a signal processing task that aims to separate signals from different sources that are mixed in the same signal, for example in a power consumption signal~\cite{FIGUEIREDO201266}. This type of problem does not correspond to ours.

For time series, \textit{temporal disaggregation} refers to methods to reconstruct a high frequency time series consistent with a lower frequency time series (the sum or average of the values of the high frequency series). It is a kind of oversampling. 
It is used in econometrics to refine annual, monthly or quarterly estimates of indicators at finer time scales. The seminal work in this field is the Chow and Lin algorithm~\cite{chow1971best} of which many variants have been proposed. 
Nevertheless, in this approach of the problem, it is necessary to have regular data about the grass growth. However, we have only one cumulative value per example in our case. The application of this type of problem is therefore likely to be inefficient. 

The disaggregation problem can also be seen as a special case of a more general problem: that of estimating individual values from aggregated data. This problem is also frequently encountered in economic applications where statistical data (for example, voting statistics) are known only for groups of people, but not individually. Mathematically, it is a matter of reconstructing a joint distribution from marginal distributions. For our application, the marginal would be the aggregate and we aim to reconstruct the distribution for the different 10-day periods. 
This type of problem can be solved for example by \textit{Iterative Proportional Fitting} (IPF) algorithms or ecological inference~\cite{king1999solution}.\footnote{The term \textit{ecological inference} does not particularly refer to ecological data.} 
For the latter approach, Quinn~\cite{quinn2004ecological} proposes a variant that takes into account temporal dependencies between values by making the assumption on dynamics. Nevertheless, the hypotheses on these dynamics can be difficult to propose, so we preferred an approach focused only on the data.

Finally, the field of machine learning has also been interested in this type of problem, notably by proposing alternatives to IPF based on Sinkhorn normal forms~\cite{idel2016review}. This formulation brings the IPF problem closer to the optimal transport problem. Nevertheless, it does not take into account the specificity of time series data.

\section{Conclusion}
We presented the temporal disaggregation of the annual cumulative grass growth. We addressed this problem by using a time series forecasting, using available exogenous climate data. 
Through our experiments, we identified a method based on the learning of a SVR model with an average RMSE lower than $12.4\pm 4.3\, kg~DM.ha^{-1}.d^{-1}$. 
The experiments showed that the initialization can be done using the average values of the first three 10-day periods without worsen the results. 
It is worth noting that the good performances of the approach is also due to available climatic variables that are known to be strongly related to the grass growth.

Finally, the experiments showed that the annual cumulative growth improves the accuracy of the model. Nevertheless, this improvement is small and the availability of reliable information on this annual cumulative growth is not guaranteed in the future. If the initial problem was to disaggregate this quantity, the results are in fact very good without using it directly.  
Thereafter, it does not seem necessary to pursue its use. 

Then, the first perspective is to better investigate the hyper-parameters of the models. More specifically, the RF overfitting may be corrected with better choices of the hyper-parameters. In addition, regularisation (Lasso or Ridge) may be evaluated.
The second perspective of this work is the investigation of time series intrinsic regression~\cite{tan2021time} as a new possible solution for our problem.
The third perspective of this work is to explore other variants. In particular, the initial choice of the period length may be changed. Reducing this period, for example to the week, will likely help the auto-regressive algorithms to be more accurate, and thus improve the disaggregation.
%
%
Finally, the selected prediction model will feed a dairy farming simulation model called FARM-AQAL developed in the framework of the European project GENTORE.

%
%

\bibliographystyle{plain}
\bibliography{biblio}

\begin{thebibliography}{10}

\bibitem{awad2015support}
Mariette Awad and Rahul Khanna.
\newblock Support vector regression.
\newblock In {\em Efficient learning machines}, pages 67--80. Springer, 2015.

\bibitem{breiman2001random}
Leo Breiman.
\newblock Random forests.
\newblock {\em Machine learning}, 45(1):5--32, 2001.

\bibitem{brisson2003overview}
Nadine Brisson, Christian Gary, Eric Justes, Romain Roche, Bruno Mary,
  Dominique Ripoche, Daniel Zimmer, Jorge Sierra, Patrick Bertuzzi, Philippe
  Burger, et~al.
\newblock An overview of the crop model {STICS}.
\newblock {\em European Journal of agronomy}, 18(3-4):309--332, 2003.

\bibitem{chow1971best}
Gregory~C Chow and An-loh Lin.
\newblock Best linear unbiased interpolation, distribution, and extrapolation
  of time series by related series.
\newblock {\em The review of Economics and Statistics}, pages 372--375, 1971.

\bibitem{deMartonne26}
Emmanuelle de~Martonne.
\newblock Ar{\'e}isme et indice d'aridit{\'e}.
\newblock {\em Comptes rendus de L'Academie des Sciences}, 182:1395--1398,
  1926.

\bibitem{FIGUEIREDO201266}
Marisa Figueiredo, Ana de~Almeida, and Bernardete Ribeiro.
\newblock Home electrical signal disaggregation for non-intrusive load
  monitoring ({NILM}) systems.
\newblock {\em Neurocomputing}, 96:66--73, 2012.

\bibitem{Graux2021:data}
Anne-Isabelle Graux.
\newblock Growth and annual valorisation of breton grasslands simulated by
  {STICS}, and associated climate.
\newblock Portail Data INRAE, 2021.

\bibitem{idel2016review}
Martin Idel.
\newblock A review of matrix scaling and sinkhorn's normal form for matrices
  and positive maps.
\newblock {\em arXiv preprint arXiv:1609.06349}, 2016.

\bibitem{king1999solution}
Gary King and John Fox.
\newblock A solution to the ecological inference problem: Reconstructing
  individual behavior from aggregate data.
\newblock {\em Canadian Journal of Sociology}, 24(1):150, 1999.

\bibitem{quinn2004ecological}
Kevin~M Quinn.
\newblock Ecological inference in the presence of temporal dependence.
\newblock In {\em Ecological Inference: New Methodological Strategies}, pages
  207--233. Cambridge University Press, 2004.

\bibitem{tan2021time}
Chang~Wei Tan, Christoph Bergmeir, Fran{\c{c}}ois Petitjean, and Geoffrey~I
  Webb.
\newblock Time series extrinsic regression.
\newblock {\em Data Mining and Knowledge Discovery}, 35(3):1032--1060, 2021.

\end{thebibliography}

\end{document}